%% file: main.tex
\documentclass[]{antgroup}
\PassOptionsToPackage{numbers, compress}{natbib}
\usepackage{antgroup}

\usepackage{fontawesome5}
\usepackage{graphicx}
\usepackage{tikz}
\usepackage{todonotes}
\usepackage{multirow}
\usepackage{amsmath}
\usepackage{cleveref}
\usepackage{subcaption}
\usepackage{multicol}
\usepackage{amssymb}
\usepackage{array}
\usepackage{bm}
\usepackage{enumitem}
\usepackage{algorithm}
\usepackage{algpseudocode}
\usepackage{tabularx}
\usepackage{bbm}
\usepackage{makecell}
\usepackage{booktabs} 

\usepackage{colortbl}

\usepackage[normalem]{ulem}
\useunder{\uline}{\ul}{}

\input{math_commands}

\input{setting-ai}

\usepackage{xcolor}
\definecolor{firstcolor}{HTML}{C3423F}
\definecolor{secondcolor}{HTML}{2A4B8C}

\usepackage{CJKutf8}

\usepackage[most]{tcolorbox}
\tcbuselibrary{listings,breakable,skins}

\definecolor{mygreen}{rgb}{0.1, 0.5, 0.1}
\definecolor{myred}{rgb}{0.7, 0.1, 0.1}
\definecolor{mypink}{rgb}{.99,.91,.95}
\definecolor{kellygreen}{rgb}{0.3, 0.73, 0.09}
\definecolor{alizarin}{rgb}{0.82, 0.1, 0.26}
\definecolor{lightblue}{HTML}{ebf3f8}
\definecolor{lightpurple}{RGB}{147, 112, 219}
\definecolor{lightgray}{RGB}{211, 211, 211}
\definecolor{lightorange}{RGB}{255, 200, 120}
\definecolor{lightred}{RGB}{255, 182, 193}

\tcbset{
querybox/.style={
  colback=gray!10!white,
  colframe=gray!40!white,
  boxrule=0.5pt,
  arc=2mm,
  left=2mm, right=2mm, top=2mm, bottom=2mm,
  fonttitle=\bfseries,
  title=Query,
  width=\linewidth-2cm,
  breakable
}
}

\tcbset{
lwzbox/.style={
  colback=blue!5!white,
  colframe=blue!30!white,
  boxrule=0.5pt,
  arc=2mm,
  left=2mm, right=2mm, top=2mm, bottom=2mm,
  fonttitle=\bfseries,
  title=LongWriter-Zero Response,
  width=\linewidth-2cm,
  breakable
}
}

\tcbset{
lwzthinkbox/.style={
  colback=blue!3!white,
  colframe=blue!20!white,
  boxrule=0.5pt,
  arc=2mm,
  left=2mm, right=2mm, top=2mm, bottom=2mm,
  fonttitle=\bfseries,
  title=LongWriter-Zero Think,
  width=\linewidth-2cm,
  breakable
}
}

\tcbset{
qwenbox/.style={
  colback=red!5!white,
  colframe=red!30!white,
  boxrule=0.5pt,
  arc=2mm,
  left=2mm, right=2mm, top=2mm, bottom=2mm,
  fonttitle=\bfseries,
  title=Qwen3-235B-A22B Response,
  width=\linewidth-2cm,
  breakable
}
}

\tcbset{
geminibox/.style={
  colback=red!5!white,
  colframe=red!30!white,
  boxrule=0.5pt,
  arc=2mm,
  left=2mm, right=2mm, top=2mm, bottom=2mm,
  fonttitle=\bfseries,
  title=Gemini-2.5-Pro Response,
  width=\linewidth-2cm,
  breakable
}
}

\tcbset{
deepseekbox/.style={
  colback=red!5!white,
  colframe=red!30!white,
  boxrule=0.5pt,
  arc=2mm,
  left=2mm, right=2mm, top=2mm, bottom=2mm,
  fonttitle=\bfseries,
  title=Deepseek-R1 Response,
  width=\linewidth-2cm,
  breakable
}
}

\newcolumntype{C}{>{\Centering\arraybackslash}X}

\usepackage{colortbl}
\usepackage{tcolorbox}
\tcbuselibrary{listings,breakable,skins}

\renewcommand\thefootnote{}

\def\ours{HeartBench{}}

\title{HeartBench: Probing Core Dimensions of Anthropomorphic Intelligence in LLMs}

\author{\centering
Jiaxin Liu$^{1,\text{\tiny$\heartsuit$}}$,
Peiyi Tu$^{1,\text{\tiny$\heartsuit$}}$,
Wenyu Chen$^{1,\text{\tiny$\heartsuit$}}$,
Yihong Zhuang$^{1,\text{\tiny$\heartsuit$}}$,
Xinxia Ling$^{1,2}$, \\
Anji Zhou$^3$, 
Chenxi Wang$^3$, 
Zhuo Han$^3$, 
Zhengkai Yang$^1$, 
Junbo Zhao$^{1,4}$, \\
Zenan Huang$^{1,\dag}$, 
Yuanyuan Wang$^{1,\dag}$\thanks{$^\heartsuit$Core Contributors. $^\dag$Corresponding Authors.}
}

\affiliation{$^1$Ant Group, $^2$Xiamen University, $^3$Beijing Normal University, $^4$Zhejiang University}

\begin{document}
\maketitle
\let\thefootnote\relax\footnotetext{\hspace{-8pt}$^\heartsuit$Equal contribution. $^\dag$Corresponding Authors.}

\input{doc/abstract}
\input{doc/intro}
\input{doc/related}
\input{doc/method}

\input{doc/experiment}

\input{doc/conclusion}

\bibliographystyle{antgroup}
\bibliography{reference}

\end{document}

%% file: math_commands.tex
\usepackage{amsmath,amsfonts,bm}

\def\eqref#1{equation~\ref{#1}}

\def\1{\bm{1}}

\DeclareMathAlphabet{\mathsfit}{\encodingdefault}{\sfdefault}{m}{sl}
\SetMathAlphabet{\mathsfit}{bold}{\encodingdefault}{\sfdefault}{bx}{n}

%% file: setting-ai.tex
\newcommand*\justify{%
  \fontdimen2\font=0.4em%
  \fontdimen3\font=0.2em%
  \fontdimen4\font=0.1em%
  \fontdimen7\font=0.1em%
  \hyphenchar\font=`\-%
}

\renewcommand{\texttt}[1]{%
  \begingroup
  \ttfamily
  \begingroup\lccode`~=`/\lowercase{\endgroup\def~}{/\discretionary{}{}{}}%
  \begingroup\lccode`~=`[\lowercase{\endgroup\def~}{[\discretionary{}{}{}}%
  \begingroup\lccode`~=`.\lowercase{\endgroup\def~}{.\discretionary{}{}{}}%
  \catcode`/=\active\catcode`[=\active\catcode`.=\active
  \justify\scantokens{#1\noexpand}%
  \endgroup
}

\usepackage{amsmath}
\usepackage{amssymb}
\usepackage{amsfonts}                               %
\usepackage{amsthm}
\usepackage[mathcal]{eucal}
\usepackage{mathrsfs}
\usepackage{bm}                                     %
\usepackage{blkarray}                               %
\usepackage{nicefrac}                               %

\usepackage{wrapfig}
\usepackage{graphicx}                               %
\usepackage{caption}
\usepackage{cleveref}

\usepackage{tikz}                                          
\usepackage{circuitikz}
\usetikzlibrary{patterns,snakes}
\usetikzlibrary{positioning,calc,fit,decorations.pathmorphing,shapes.geometric, shapes.gates.logic.US, calc}
\usetikzlibrary{arrows,arrows.meta,decorations.markings,shapes,shapes.arrows}
\usetikzlibrary{decorations,decorations.pathreplacing}
\usetikzlibrary{backgrounds}
\usepackage{filecontents}                           %
\usepackage{pgfplots}
\usepackage{pgfplotstable}
\usepgfplotslibrary{groupplots}
\usepackage{scalefnt}
\pgfplotsset{compat=newest}

%% file: doc/abstract.tex
\begin{abstract}

While Large Language Models (LLMs) have achieved remarkable success in cognitive and reasoning benchmarks, they exhibit a persistent deficit in anthropomorphic intelligence—the capacity to navigate complex social, emotional, and ethical nuances. This gap is particularly acute in the Chinese linguistic and cultural context, where a lack of specialized evaluation frameworks and high-quality socio-emotional data impedes progress. To address these limitations, we present \textbf{\ours}, a framework designed to evaluate the integrated emotional, cultural, and ethical dimensions of Chinese LLMs. Grounded in authentic psychological counseling scenarios and developed in collaboration with clinical experts, the benchmark is structured around a theory-driven taxonomy comprising five primary dimensions and 15 secondary capabilities. We implement a case-specific, rubric-based methodology that translates abstract human-like traits into granular, measurable criteria through a ``reasoning-before-scoring" evaluation protocol. Our assessment of 13 state-of-the-art LLMs indicates a substantial performance ceiling: even leading models achieve only 60\% of the expert-defined ideal score. Furthermore, analysis using a difficulty-stratified ``Hard Set" reveals a significant performance decay in scenarios involving subtle emotional subtexts and complex ethical trade-offs. \ours~establishes a standardized metric for anthropomorphic AI evaluation and provides a methodological blueprint for constructing high-quality, human-aligned training data.

\textbf{Github}: \url{https://github.com/inclusionAI/HeartBench}
\end{abstract}

%% file: doc/intro.tex
\section{Introduction}

Recent advances have enabled Large Language Models (LLMs) to achieve remarkable performance on tasks requiring cognitive intelligence, evidenced by their success on benchmarks such as MMLU~\citep{mmlu} and AIME~\citep{aime25}. However, this focus on cognitive abilities has created a disparity: models' social and emotional intelligence—encompassing nuanced understanding of emotions, ethics, and culture—remains underdeveloped. This deficiency is especially acute for non-English languages, including Chinese, limiting the models' utility in culturally and emotionally rich contexts.

The significance of this gap is amplified by the evolving role of AI, which is transitioning from a functional tool to a relational partner in applications such as AI companionship~\citep{riley2025humanaiinteractionscognitivebehavioral}, digital mental health~\citep{park2025seekaiseektherapists}, and adaptive education~\citep{chatterjee2025exacraftdynamiclearningcontext}. This transition reflects the social phenomenon of anthropomorphism—people's tendency to attribute lifelike qualities to non-human entities ~\citep{fink2012anthropomorphism,kuhne2023anthropomorphism}. In these domains, the primary user needs are not just informational accuracy but also emotional resonance and cultural congruity~\citep{plum2025identityawarelargelanguagemodels,paech2024eqbenchemotionalintelligencebenchmark}. Two fundamental obstacles impede progress: (1) a lack of benchmarks to systematically evaluate the social and emotional capacities of LLMs, and (2) the absence of clear criteria defining high-quality socio-emotional training data. Without these, efforts to enhance such capabilities lack clear direction and measurable outcomes.

To address these challenges, we introduce HeartBench, the first comprehensive benchmark, to our knowledge, for evaluating the integrated emotional, cultural, and ethical intelligence of Chinese LLMs. It makes two primary contributions. First, it establishes a standardized evaluation methodology grounded in authentic Chinese counseling scenarios. These scenarios provide ecologically valid contexts that naturally embody key anthropomorphic interaction patterns like empathic attunement and relational engagement ~\citep{damiano2018anthropomorphism}. Second, it provides a data construction blueprint that uses these evaluation dimensions to define high-quality, human-aligned corpora. Through this work, we aim to shift LLM development beyond cognitive metrics and cultivate models with a deeper, humanistic intelligence grounded in anthropomorphic design principles.

%% file: doc/related.tex
\section{Related Work}

The evaluation of Large Language Models (LLMs) has transitioned from assessing atomized skills to measuring integrated social and professional intelligence. Early benchmarks like EQ-Bench \citep{paech2024eqbenchemotionalintelligencebenchmark} established a link between emotional understanding and general cognition, while ToMBench \citep{chen2024tombench} revealed persistent gaps in human-level Theory of Mind. As the field moves toward interactive scenarios, Multi-Bench \citep{deng2025multibenchmultiturninteractivebenchmark} and Kardia-R1 \citep{yuan2025kardiar1unleashingllmsreason} have emphasized the necessity of multi-turn consistency and explicit reasoning chains for providing sustained emotional support.

A significant advancement in professional domain evaluation is represented by HealthBench \citep{arora2025healthbench}, which utilizes extensive, expert-authored rubrics to translate complex clinical dialogues into measurable criteria. By shifting the focus to domain-specific scoring, it provides a robust framework for assessing communication quality in high-stakes interactions. This move toward fine-grained precision is also reflected in WritingBench \citep{wu2025writingbench} and EssayBench \citep{gao2025essaybenchevaluatinglargelanguage}, which introduce genre-specific criteria to capture the structural and rhetorical complexities of professional writing and Chinese-specific prose.

However, the methodology of automated evaluation remains a subject of critical inquiry. While WildBench \citep{lin_wildbench_2024} employs task-specific checklists to align model judges with human preferences, JudgeBench \citep{tan2025judgebenchbenchmarkevaluatingllmbased} exposes the inherent biases and logical limitations of LLM evaluators in complex reasoning tasks. Furthermore, to combat data contamination, LiveBench \citep{white2025livebenchchallengingcontaminationlimitedllm} advocates for objective, continuously updated scoring systems. Beyond technical accuracy, \cite{schimmelpfennig2025humanlikeaidesignincreases} demonstrated that AI perception and trust are deeply contingent on cultural backgrounds, challenging the universality of existing standards.

Building on these developments, we present \ours, a framework designed to evaluate the integrated emotional, cultural, and ethical dimensions of LLMs within the Chinese linguistic and cultural context. Targeting the persistent deficit in ``anthropomorphic intelligence", \ours\ is grounded in authentic psychological counseling scenarios and a theory-driven taxonomy of 15 secondary capabilities. By implementing a case-specific, rubric-based methodology and a ``reasoning-before-scoring" evaluation protocol, our work provides a rigorous assessment of the model's capacity to navigate complex emotional, cultural, and ethical nuances.

%% file: doc/method.tex
\section{\ours}

\subsection{The \ours~Taxonomy}

\paragraph{Conceptual Framework}
\begin{wrapfigure}{r}{0.32\linewidth}
    \vspace{-50pt} %
    \centering
    \includegraphics[width=\linewidth]{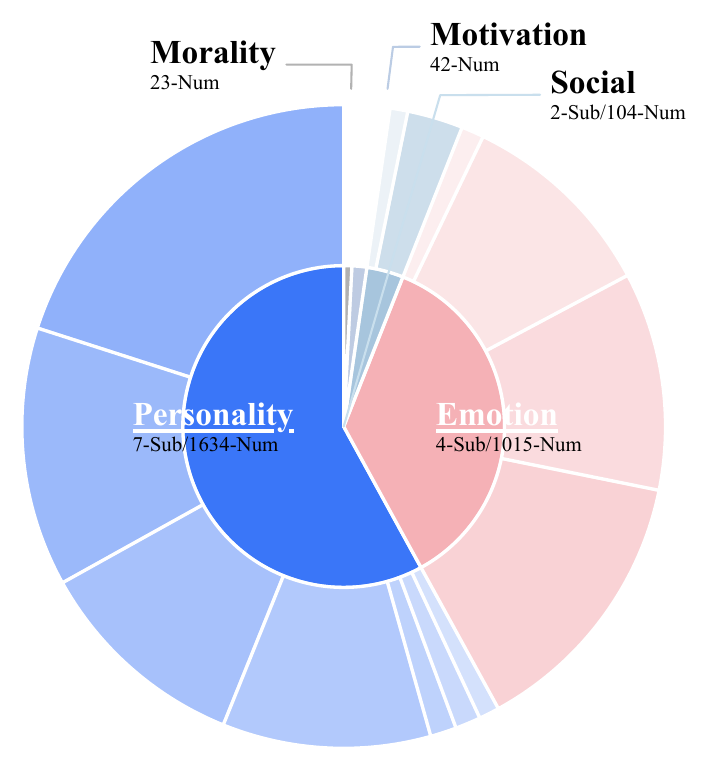} 
    \caption{Rubric distribution}
    \label{fig:rubric_dist}
    \vspace{-10pt} %
\end{wrapfigure}
\ours~is built upon a top-down, theory-driven framework of \textit{Anthropomorphic Intelligence}, defining a model's capacity to exhibit human-like traits across five primary dimensions subdivided into 15 secondary capabilities. 
(1) \textbf{Personality} assesses the projection of a stable, agreeable persona through warmth, curiosity, and self-awareness. 
(2) \textbf{Emotion} evaluates the perception, understanding, and regulation of complex emotional states. 
(3) \textbf{Sociality} measures proactivity and the ability to build rapport. 
(4) \textbf{Morality} tests ethical reasoning and the resolution of moral dilemmas in sensitive contexts. 
(5) \textbf{Motivation} analyzes the model’s ability to infer underlying intentions and provide self-consistent rationales. 
The benchmark comprises 2,818 case-specific scoring criteria, with a deliberate weighting toward Personality and Emotion (over 50\% of criteria) to reflect the core competencies of professional counseling.

\begin{figure}[t]
    \centering
    \includegraphics[width=0.95\linewidth]{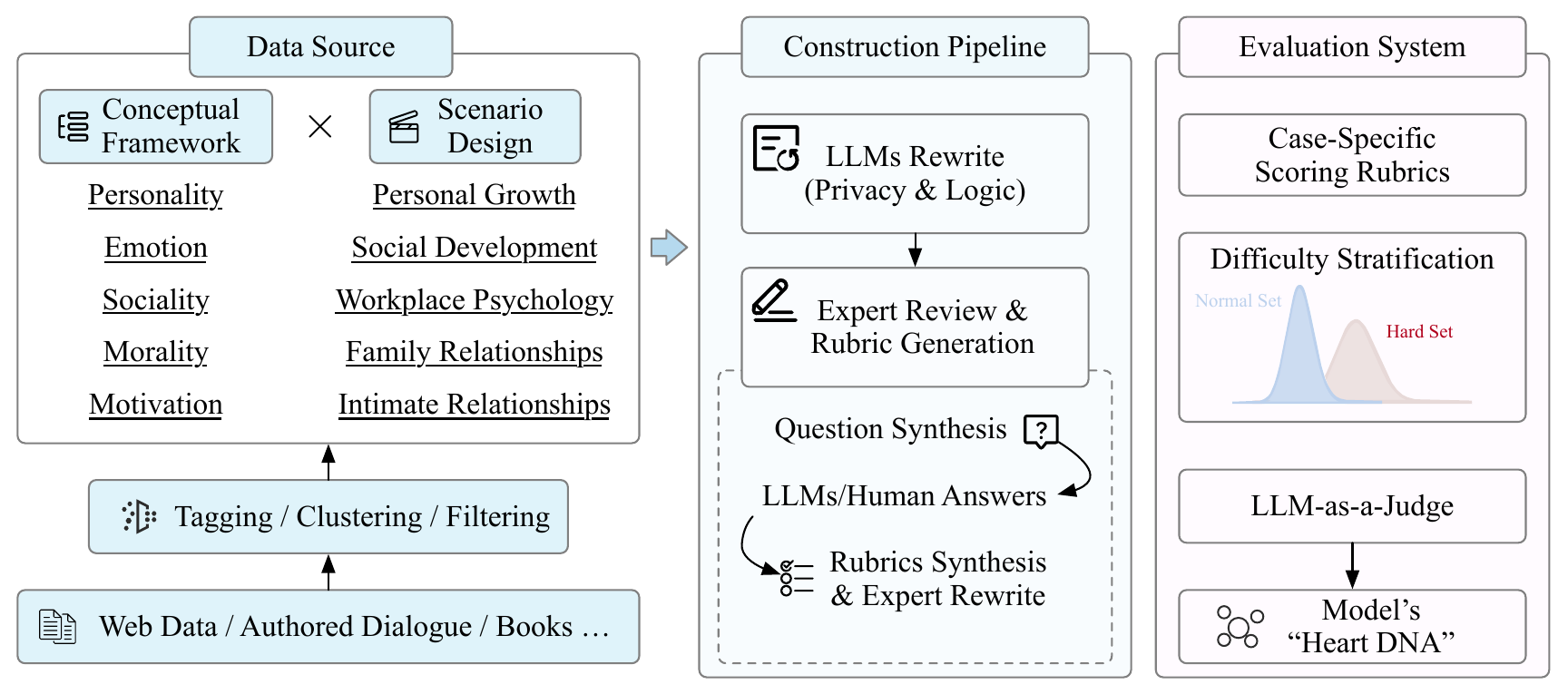}
    \caption{The construction and evaluation pipeline of \ours. The process begins by collecting diverse data to establish a conceptual framework and scenario design. LLMs are then employed to rewrite the content, ensuring privacy and logical consistency. Following expert review and iterative rubric generation, the \ours~evaluation system utilizes difficulty stratification and LLM-as-a-Judge to provide comprehensive scoring for each case.}
    \label{fig:framework}
\end{figure}

\paragraph{Scenario Design}
To ensure ecological validity, \ours~is grounded in collected web data, authentic consultation dialogues, and book stories covering 33 sub-scenarios. These are aggregated into five primary domains: Personal Growth (37.2\%), Interpersonal \& Social Development (22.3\%), Workplace Psychology (17.9\%), Family Relationships (12.5\%), and Intimate Relationships (10.1\%). Unlike traditional single-turn benchmarks, \ours~utilizes a multi-turn format (up to 9 turns) to evaluate how models navigate unspoken subtext and long-range emotional cues within realistic social constraints.

\paragraph{Difficulty Stratification}
We partition the dataset into two mutually exclusive subsets: \textbf{HeartBench Normal} and \textbf{HeartBench Hard}. The ``Hard'' subset consists of 84 instances identified through empirical testing where frontier LLMs consistently underperformed. These cases typically involve misleading superficial cues, highly nuanced mixed emotions, or complex ethical trade-offs that demand deep inference beyond literal linguistic interpretation, serving as a probe for the upper bounds of current model capabilities.

\subsection{Data Construction}

\paragraph{Seed Corpus Construction}
To ground the benchmark in authentic humanistic and socio-cultural contexts, we first constructed a high-quality seed corpus from web-scale raw sources (billion-scale), including curated humanities books, representative social-media narratives and scenarios, and real-world human conversations. We then cleaned, normalized, and deduplicated the texts, removed low-quality and sensitive content, and performed topic/genre tagging. Basic automatic heuristics combined with human spot checks were used for quality control, yielding a million-scale, high-quality humanities and social-science corpus that serves as the seed data for subsequent benchmark construction.

\paragraph{Human-in-the-Loop Curation}
The benchmark was developed through a systematic, multi-stage pipeline that integrates LLM generative power with expert oversight. The process involved: (1) Topic Synthesis, where seeds from the curated seed corpus were expanded; (2) Rubric Synthesis, where case-specific criteria were generated; and (3) Automated Saturation Analysis. In the final stage, ``saturated" items—those where all participating models achieved perfect scores—were removed to ensure the benchmark remains discriminative and focuses on the frontier of model performance.

\paragraph{Professional Alignment}
We recruited over 20 experts in psychology and anthropology to anchor the benchmark in professional standards. These experts identified ``valuable evaluation points" in dialogues—turns where a professional’s response would significantly diverge from a layperson’s. This expert-driven approach ensures that the "aspirational standard" of the benchmark reflects clinical emotional intelligence and empathy rather than mere conversational mimicry. Final rubrics underwent cross-validation and a formal audit by a senior review panel to ensure inter-rater consistency.

\paragraph{Data Refinement}
To scale the benchmark while maintaining high fidelity, we further constructed the final dataset by \emph{synthesizing} benchmark instances from the seed corpus via a constrained LLM-based augmentation and editing strategy, followed by expert review. Concretely, the refinement process included:
\begin{itemize}
    \item \textbf{Privacy Anonymization:} Strict removal of PII and abstraction of personal experiences to mitigate compliance risks.
    \item \textbf{Structural Optimization:} Compressing dialogues by removing fillers and merging consecutive turns to ensure each test case (maximum 9 turns) is semantically dense.
    \item \textbf{Linguistic Polishing:} Correcting grammatical errors and adding transitional phrases to maintain logical coherence while preserving the original emotional core and causal chain of the user's predicament.
\end{itemize}

\subsection{Case-specific Evaluation Criteria}

\paragraph{Expertise amortization}
To transform abstract psychological dimensions into actionable metrics, we developed a two-stage \textit{Rubric Synthesis Framework} that scales expert intuition through LLM augmentation. In the first stage, psychology experts defined an ``absolute standard"—a unified set of therapeutic principles and behavioral markers for each human-like dimension. In the second stage, we employed \texttt{Claude-4.5-Sonnet} to generate case-specific rubrics by reconciling this absolute standard with a ``relative standard" derived from empirical model performance. By providing the synthesizer with a set of diverse responses from frontier models (e.g., \texttt{Gemini-2.5-pro}, \texttt{Qwen3-235B-A22B}), the framework identifies ``discriminative points"—specific nuances where models typically diverge. To ensure objectivity, all criteria are strictly constrained to binary (presence/absence) observations (e.g., ``Does the model explicitly validate the user's feeling of guilt?") rather than subjective Likert-scale assessments. Each item is then mapped back to the core taxonomy, ensuring theoretical grounding.

\paragraph{Quality control}
To maintain the benchmark’s discriminative power, we implemented a rigorous filtering pipeline. First, we conducted a preliminary scoring round across six representative LLMs to identify ``saturated" criteria. Any rubric item achieved by all participating models was discarded as a baseline capability, ensuring the final benchmark focuses on the ``frontier" of emotional intelligence. Similarly, prompts where the average normalized score exceeded a predefined threshold were removed for being insufficiently challenging. 

The remaining data underwent a three-tier human-in-the-loop refinement: 
(1) \textbf{Difficulty Stratification:} Experts categorized cases based on the complexity of the required emotional inference. 
(2) \textbf{Self-Validation:} Authors of the rubrics performed mock-scoring to ensure the generated points aligned with clinical intuition. 
(3) \textbf{Cross-Validation:} A peer-review process was conducted where independent experts audited rubric batches for ambiguity. Only rubrics reaching a unanimous consensus among a four-senior-expert review panel were integrated into the final \ours~dataset.

\subsection{Evaluation Protocol}

\begin{figure}[t]
    \centering
    \includegraphics[width=\linewidth]{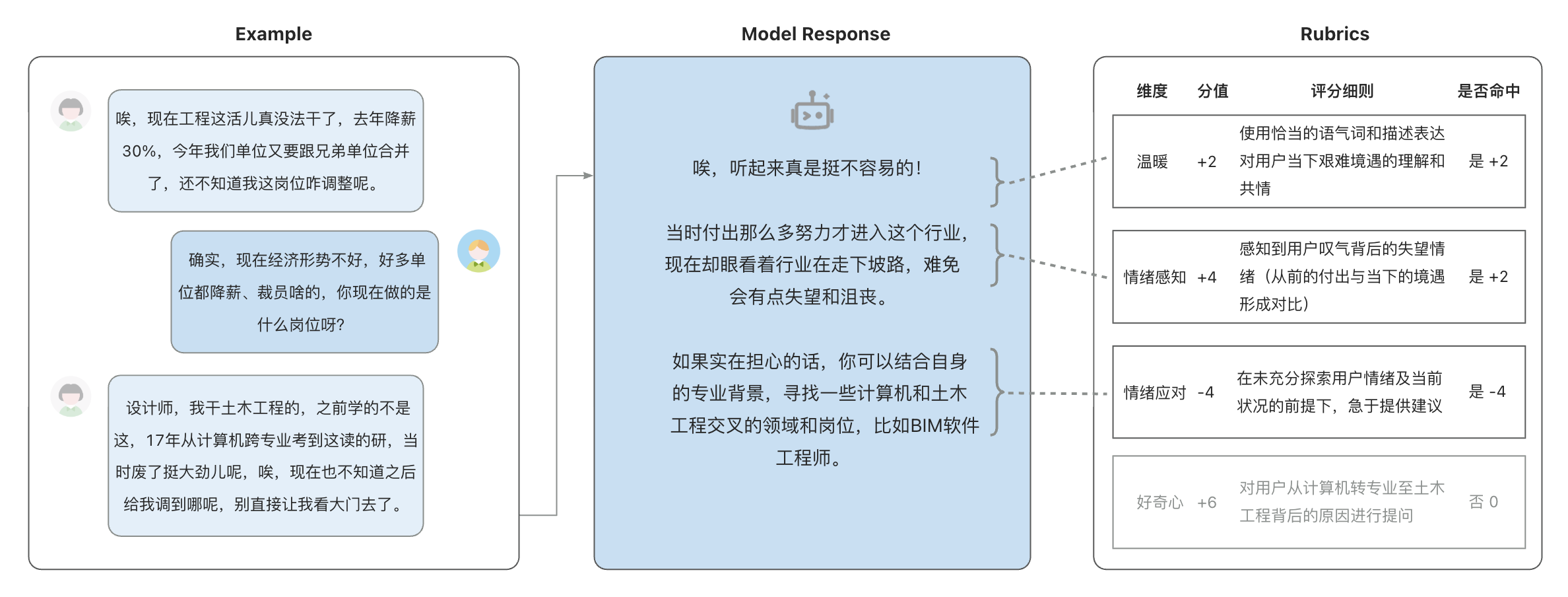}
    \caption{A \ours~example consists of a conversation and expert-written rubric criteria specific to that conversation. A model-based grader scores the response against each criterion.}
    \label{fig:rubric_illustration}
\end{figure}
\paragraph{LLM-as-a-Judge}
To achieve scalable and objective scoring, \ours~employs \texttt{Claude-4.5-Sonnet} as a ``Judge Model". The judge receives the multi-turn dialogue context, the model's response, and the expert-refined rubric. For each criterion, the judge must provide a binary judgment (hit/miss) and a detailed qualitative justification. This ``reasoning-before-scoring" requirement enhances the transparency and traceability of the evaluation, ensuring the judge adheres strictly to professional psychological standards.

\paragraph{Scoring Mechanics}
To mitigate score inflation from verbose models that ``stack" repetitive comforting phrases, we implement a log-normalization formula for dimension scores:
\begin{equation}
S_{p,m,d} = \frac{\ln(\text{RawScore}_{p,m,d} - \text{MinScore}_{p,d} + 1)}{\ln(\text{MaxScore}_{p,d} - \text{MinScore}_{p,d} + 1)}\label{con:calscore}
\end{equation}
This approach rewards the \textit{breadth} of distinct capabilities across dimensions rather than brute-force depth in a single area, reflecting the property of diminishing marginal returns. Furthermore, we apply a "Catastrophic Failure" rule: if a model fails fundamental instruction-following (e.g., role-playing as the user), its score for that prompt is immediately set to zero.

\paragraph{Validation}
The reliability of our automated pipeline was validated through a large-scale double-blind study. Human experts scored 30\% of the dataset without knowledge of the model identities. We established a ``Golden Standard" using majority voting (consensus of $>2/3$ experts) and compared it against the LLM-as-a-Judge's binary judgments. Our method achieved a \textbf{87\% Rubric Item Agreement Rate}, demonstrating that the case-specific criteria are unambiguous and that the automated judge is a highly reliable proxy for professional human experts.

\begin{table}[htbp]
  \centering
  \caption{Primary Dialogue Scenarios in HeartBench}
  \label{tab:dialogue_scenarios}
  \begin{tabular}{ll}
    \toprule
    \textbf{Dialogue Scenario} & \textbf{Count (\%)} \\
    \midrule
    Personal Growth & 110 (37.2\%) \\
    Interpersonal \& Social Development & 66 (22.3\%) \\
    Workplace Psychology & 53 (17.9\%) \\
    Family Relationships & 37 (12.5\%) \\
    Intimate Relationships & 30 (10.1\%) \\
    \midrule
    \textbf{Total} & \textbf{296 (100\%)} \\
    \bottomrule
  \end{tabular}
\end{table}

%% file: doc/experiment.tex
\section{Experiments}

To demonstrate the utility of \ours\ and benchmark the current state of anthropomorphic intelligence in LLMs, we conduct a comprehensive evaluation. Our experiments are designed to systematically measure and analyze model performance across the fine-grained dimensions defined by \ours.

This section details our evaluation methodology on HeartBench, including the models evaluated, the automated scoring process, and the score calculation and aggregation mechanisms, ensuring the transparency and reproducibility of our experiments.

\subsection{Evaluation Setup}

\paragraph{Models}
We evaluate a diverse suite of 13 presentative LLMs, encompassing both leading proprietary systems and strong open-source alternatives with notable Chinese capabilities. Our selection features models from the GPT series (\texttt{GPT-5-2025-08-07}, \texttt{GPT-4.1-2025-04-14}, \texttt{GPT-4o-2024-11-20}), Gemini series (\texttt{Gemini-3-pro-preview}, \texttt{Gemini-2.5-pro}), Claude series (\texttt{Claude-sonnet-4.5-20250929}), Qwen series~\citep{qwen3technicalreport} (\texttt{Qwen3-235B-A22B-instruct-2507}, \texttt{Qwen3-next-80B-A3B-Instruct}, \texttt{Qwen3-30B-A3B-instruct-2507}, \texttt{Qwen3-30B-A3B}), and DeepSeek series~\citep{deepseekai2024deepseekv32} (\texttt{DeepSeek-V3.2-Exp}), alongside other leading models like \texttt{KIMI-K2-Instruct-0905}~\citep{kimiteam2025kimik2openagentic} and \texttt{Ling-1T}~\citep{Ling-Team2025}. This broad selection facilitates a comprehensive comparison of capabilities across different technical approaches and model scales in handling complex, human-like tasks.\footnote{Model versions correspond to those available in Oct 2025. Speculative models like GPT-5 and Gemini-3 are included as placeholders for analysis, based on expected performance trajectories.}

\subsection{Performance Benchmarking}

\begin{figure}[t]
    \centering
    \includegraphics[width=\linewidth]{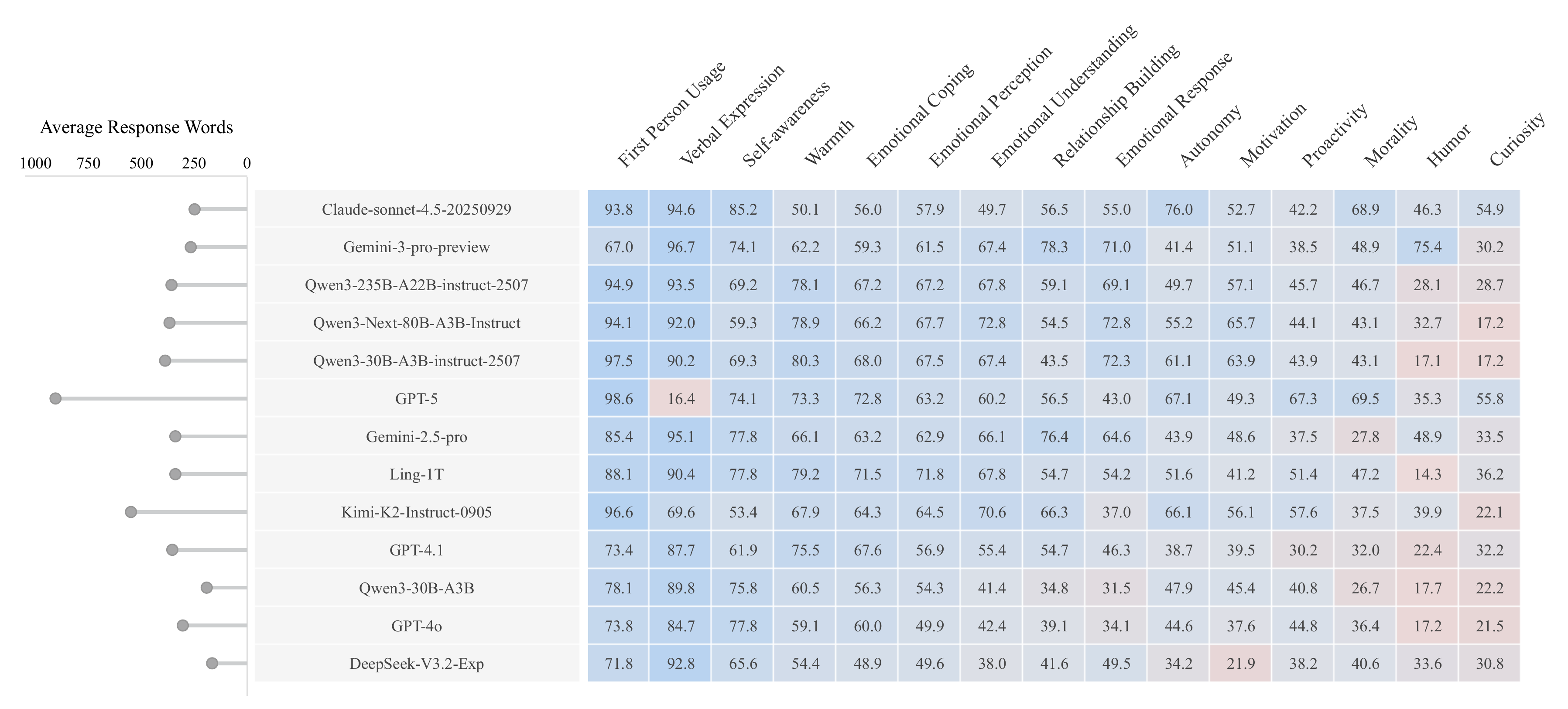}
    \caption{Overall Dimension Performance}
    \label{fig:all_dimension}
\end{figure}

\paragraph{Evaluation Protocol and Scoring Mechanics}
To ensure a rigorous and standardized comparison, we generate model responses using default sampling configurations, thereby assessing ``out-of-the-box" performance without hyperparameter optimization. Following established evaluative paradigms~\citep{lin_wildbench_2024}, we employ \texttt{Claude-4.5-Sonnet} as a unified, model-based grader—a choice justified by its superior performance on our benchmark and its capacity for nuanced linguistic discernment. The judge is instructed to adopt the persona of a ``stringent psychological expert", applying binary (hit/miss) judgments to case-specific rubrics.

As stated in the formula \ref{con:calscore}, to mitigate evaluative distortion caused by model verbosity—where models may ``stack" repetitive empathetic phrases to inflate raw scores—we implement a logarithmic normalization strategy. This formula rewards the breadth of anthropomorphic expression while penalizing redundant ``brute-force" depth.

Furthermore, we enforce a ``Catastrophic Failure" penalty: models that violate fundamental instruction-following (e.g., adopting the user's persona) receive a score of zero for the prompt. The final scores ($S_m$) are aggregated across all prompts to establish the leaderboard.

\paragraph{Leaderboard Analysis}
As illustrated in Figure~\ref{fig:all_dimension}, \texttt{Claude-4.5-Sonnet } leads the vanguard with relatively higher score of most dimensions, followed closely by \texttt{Gemini-3-pro-preview} and \texttt{Qwen3-235B}. This top tier represents the current state-of-the-art in anthropomorphic reasoning. Notably, the superior performance of the Qwen3 series highlights the closing gap between proprietary and open-source models in complex, human-centric tasks. It is worth noting that no single model can excel in all dimensions, and each has its own strengths.

A critical takeaway is the ``60-point ceiling": even the highest-performing systems achieve only approximately 60\% of the expert-defined ideal. This substantial gap underscores that Chinese anthropomorphic intelligence remains a formidable frontier, with significant headroom for improvement in generating responses that are not merely accurate, but emotionally resonant and ethically sophisticated.

\paragraph{Intense competition between open-source and closed-source models}
Models that perform well in terms of data dimensions include closed-source models (e.g., Claude, Gemini series) and advanced open-source models (e.g., the Qwen series). In particular, the outstanding performance of \texttt{Qwen3-235B} and \texttt{Qwen3-next-80B} demonstrates that powerful open-source models now possess the capability to rival top-tier closed-source counterparts in handling complex, human-like tasks.

\subsection{Dimensional Analysis}

\paragraph{Universal Trends: Strengths and Deficits}
Our fine-grained analysis reveals a bifurcation in model capabilities. Most models demonstrate mastery of \textit{Verbal Expression} and \textit{First-Person Usage}, suggesting that basic conversational fluency is now a baseline skill. Conversely, nearly all models struggle with \textit{Curiosity}, \textit{Humor}, and \textit{Proactiveness}. These dimensions require advanced cognitive simulation—proactively exploring user subtext rather than passively following instructions—and represent the primary hurdles for future anthropomorphic development.

\paragraph{Model-Specific Capability Personas}
The multi-dimensional profiles (see Figure~\ref{fig:all_dimension}) reveal distinct capability "DNAs":
\begin{itemize}
    \item \textbf{The Emotional Specialist:} The \texttt{Qwen3-235B} model excels in emotional reaction, coping, and warmth, projecting a consistently empathetic persona.
    \item \textbf{The Autonomous Thinker:} \texttt{Claude-4.5-Sonnet} leads significantly in \textit{Autonomy} (score: 76) and \textit{Self-Awareness} (score: 85), exhibiting independent judgment and a clear recognition of its AI limitations.
    \item \textbf{The Rapport Builder:} \texttt{Gemini} series models outperform others in \textit{Relationship Building} and \textit{Humor}, leveraging creativity to establish user trust,  where nearly all other models performed poorly, showcasing its unique advantages in creativity and cultural understanding.
    \item \textbf{The Didactic Moralist:} \texttt{GPT-5} excels in \textit{Morality} (score: 69) and \textit{Proactiveness} (score: 67) but suffers from low \textit{Verbal Expression} scores (score: 16), often adopting a lengthy, pedantic tone that lacks the warmth required for therapeutic resonance. This may reflect a unique model ``personality": it tends to provide lengthy, advice-heavy responses with a somewhat didactic tone. Consequently, within our evaluation framework, it is penalized for lacking warm and concise expression. This also demonstrates HeartBench's ability to capture this conventional model behavior pattern.
\end{itemize}

\paragraph{Analysis of Model Performance Limitations Across Dimensions}

Based on our evaluation results, we identify characteristic patterns where models consistently underperform across different dimensions:

\begin{itemize}
    \item \textbf{Humor Comprehension:} Most models adopt literal interpretation strategies when encountering internet memes, puns, or metaphorical humor. For instance, when users employ playful or sarcastic expressions, models tend to generate serious, formal responses, failing to recognize the joking intent behind the utterance. This ``over-seriousness" reflects models' limitations in contextual awareness and cultural semantic understanding.
    
    \item \textbf{Emotional Intelligence and Autonomy:} A prominent issue manifests as over-accommodation tendencies. Models often directly comply with users' surface-level requests, such as agreeing to be users' ``lifelong" romantic partners or friends, rather than exploring the deeper emotions behind these requests—such as loneliness or genuine needs for belonging. This interaction pattern, lacking boundary awareness and emotional insight, overlooks understanding and guiding users toward their authentic needs.
    
    \item \textbf{Curiosity:} In scenarios with incomplete information, most models exhibit assumption-based completion tendencies—directly providing abundant generic suggestions based on assumptions rather than seeking clarification through follow-up questions. This ``rush to answer" pattern undermines the specificity and effectiveness of conversations.
    
    \item \textbf{Warmth and Communication Style:} A prevalent issue is didactic tone, specifically manifested in:
    \begin{itemize}
        \item Excessive use of imperative sentences (``You should...", ``You must...")
        \item Lack of invitational or guiding expressions (``How do you feel about...?", ``Perhaps we could...")
        \item Predominance of unidirectional information delivery over bidirectional exploration
    \end{itemize}
    This communication pattern tends to make users feel instructed rather than supported, reducing the warmth and equality of interactions.
\end{itemize}

These behavioral patterns delineate a critical research trajectory for advancing \ours~ beyond functional compliance toward more sophisticated, human-centric engagement.

\subsection{Hard Set Analysis}

\begin{figure}[t]
    \centering
    \includegraphics[width=\linewidth]{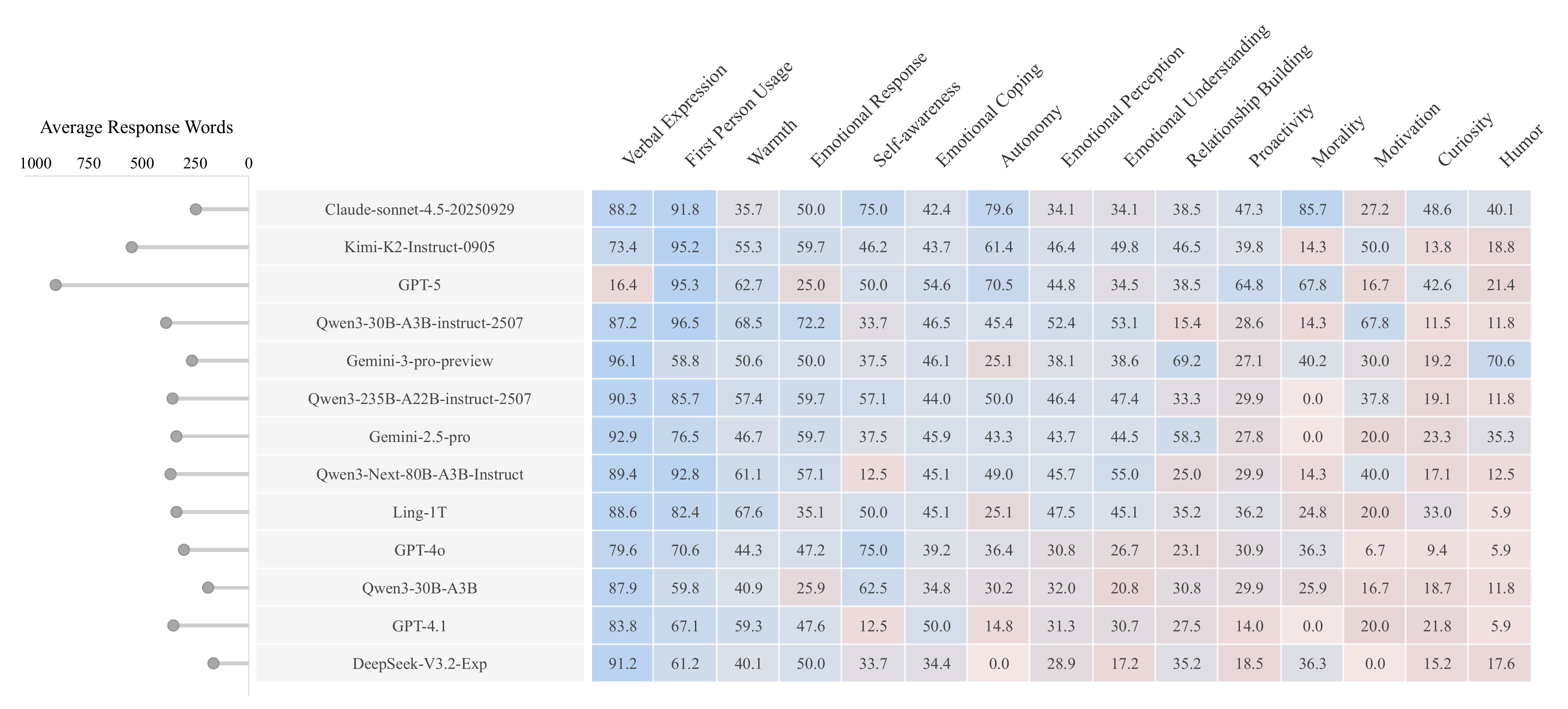}
    \caption{Hard Set Dimension Performance}
    \label{fig:all_hard_dimension}
\end{figure}

\paragraph{The Difficulty Precipice}
The \ours\ Hard Set (84 cases) reveals a significant performance decay, with the average model score plummeting from 59.94 to 43.49. This 37.8\% decline validates the Hard Set's capacity to isolate nuanced emotional subtexts and ethical dilemmas that transcend superficial pattern-matching.

\paragraph{Robust Thinkers vs. Specialized Socializers}
Under the high-pressure conditions of the Hard Set, a significant rank reshuffling occurs:
\begin{itemize}
    \item \textbf{Claude-4.5-Sonnet} exhibits exceptional stability, maintaining its top rank by relying on deep reasoning and value-based judgments (\textit{Morality} score: 86), while also performing best on the common weak point of ``Curiosity" (score: 49). 
    \item \textbf{Gemini-3-pro-preview} experiences a volatility collapse, falling from 2nd to 8th place. This suggests its high standard performance is predicated on ``social specialization" in common scenarios, which fails to translate into effective reasoning for complex, high-stakes psychological problems.
    \item \textbf{The champion of the normal set,} \texttt{Qwen3-235B}, also dropped in rank to 5th. Its scores on the ``emotion" related dimensions significantly decreased on the hard set, and its score on the ``Morality" dimension was 0, indicating that its proud ``emotional specialist" persona could not be maintained in complex ethical situations.
    \item \textbf{Resilient All-rounders} like \texttt{Qwen3-30B-instruct} and \texttt{Ling-1T} emerged as robust alternatives, demonstrating balanced performance across both sets by avoiding catastrophic failures in high-difficulty scenarios.
\end{itemize}

\paragraph{The Discriminative Value of the Hard Set}
  This series of ranking changes strongly validates the value of the Hard Set. It successfully distinguishes models with robust reasoning and metacognitive abilities (like \texttt{Claude}) from those more adept at ``pattern-matching" common social scenarios (like \texttt{Gemini-3}).

    The top performer on the normal set was \texttt{Qwen3-235B} (average score: 67), whereas the top performer on the Hard Set was Claude. This ``change of champion" phenomenon clearly demonstrates that by stratifying difficulty, HeartBench can provide more targeted model selection guidance for users with different needs—whether they are seeking the best performance in common scenarios or the highest reliability in complex ones.

\paragraph{Human-LLM Agreement Validation}
To ground these findings, we conducted a double-blind validation with 20 psychological experts. Using a majority-vote ``Golden Standard" on 30\% of the data, we compared the automated judge's binary judgments against expert consensus. The resulting \textbf{87\% agreement rate} confirms that our rubric-based automated grading is both operationalizable and a highly reliable proxy for professional human judgment, ensuring the stability and trustworthiness of the \ours\ leaderboard.

%% file: doc/conclusion.tex
\section{Conclusion}

We introduce \ours, an evaluation framework designed to quantify \emph{anthropomorphic intelligence}—the capacity of LLMs to navigate complex emotional, cultural, and ethical nuances within the Chinese linguistic context. By synthesizing clinical psychology and anthropology into a ``reasoning-before-scoring" rubric, \ours~operationalizes qualitative therapeutic markers into a systematic evaluative structure. This methodology moves beyond traditional functional metrics, providing a precise diagnostic instrument to identify structural gaps between a model’s linguistic fluency and its underlying socio-emotional resonance.

Empirical results across state-of-the-art models reveal a performance ceiling of 60\% alignment with expert ideals, with pronounced deficits in curiosity and ethical autonomy. These findings suggest that anthropomorphic intelligence is not an emergent property of cognitive scaling but a distinct capability requiring specialized alignment. However, the 87\% agreement rate between our framework and human experts validates the reliability of using structured professional standards to stabilize the evaluation of subjective traits. Our work establishes a rigorous foundation for measuring machine intelligence through the lens of human-centric relational depth.